\documentclass[10pt]{article}

\usepackage[accepted]{rlc}

\usepackage{graphicx}
\usepackage[utf8]{inputenc} % allow utf-8 input
\usepackage[T1]{fontenc}    % use 8-bit T1 fonts
\usepackage{hyperref}       % hyperlinks
\usepackage{url}            % simple URL typesetting
\usepackage{booktabs}       % professional-quality tables
\usepackage{amsfonts}       % blackboard math symbols
\usepackage{nicefrac}       % compact symbols for 1/2, etc.
\usepackage{microtype}      % microtypography
\usepackage{xcolor}         % colors

\usepackage{multirow}

\usepackage{comment}
\usepackage{float}
\usepackage{bm,amsmath}
\usepackage[inline]{enumitem}
\usepackage{subcaption}
\usepackage{mathtools}
\usepackage{natbib}
\usepackage{amssymb}

\usepackage{listings}

\usepackage{wrapfig}

\usepackage{titlesec}
\titlespacing\section{0pt}{4 pt plus 4pt minus 2pt}{2pt plus 2pt minus 2pt}

\def\Re{\mathbb{R}}

\def\Nat{{\rm I\kern\pIR N}}
\def\argmax{\mathop{\rm arg\,max}}

\newcommand{\EE}[1]{\exptE\left[#1\right]}

\def\E{{\mathcal{E}}}

\def\vec0{{\boldsymbol{0}}}

\newcommand{\eproof}{$\null\hfill\square$}
\newcommand{\beq}{\begin{equation}}
\newcommand{\eeq}{\end{equation}}
\newcommand{\beqa}{\begin{eqnarray}}
\newcommand{\eeqa}{\end{eqnarray}}
\newcommand{\beqan}{\begin{eqnarray*}}
\newcommand{\eeqan}{\end{eqnarray*}}
\newcommand{\ben}{\begin{eqnarray*}}
\newcommand{\een}{\end{eqnarray*}}

\renewcommand{\EE}[2]{\mathbb{E}_{#1\!\!}[#2]}

\newcommand{\CEE}[3]{\EE{#1}{{#2}~\middle\vert~{#3}}}
\renewcommand{\CEE}[3]{\EE{#1}{{#2}\mid{#3}}}
\def\CE#1#2{\CEE{\,}{#1}{#2}}

\def\E#1{\EE{\,}{#1}}

\def\Pr#1{{\mathbb{P}\!(#1)}}
\def\CP#1#2{\Pr{#1\mid#2}}

\def\shb{Cross-environment Hyperparameter Setting Benchmark}
\def\shbabb{CHS}
\def\hyper{hyperparameter}
\def\hypers{hyperparameters}
\def\Hyper{Hyperparameter}
\def\Hypers{Hyperparameters}

\def\perf{g}

\def\normalizeE#1{N_{\env}(#1)}

\def\Perf{G}
\def\Envs{\mathcal{E}}
\def\env{E}
\def\Params{\Theta}
\def\param{\theta}
\def\shbparam{\param_{\text{\shbabb{}}}}

\newif\ifcomments

\commentsfalse     % Disable comments

\title{The Cross-environment Hyperparameter Setting Benchmark for Reinforcement Learning}

\author{%
  Andrew Patterson, Samuel Neumann, Raksha Kumaraswamy, Martha White, Adam White \\
  Department of Computing Science, University of Alberta\\
  \{\texttt{ap3,sfneuman,kumarasw,whitem,amw8}\}\texttt{@ualberta.ca}
}

\begin{document}

\maketitle

\begin{abstract}
  This paper introduces a new empirical methodology, the \shb{}, that compares RL algorithms across environments using a single \hyper{} setting, encouraging algorithmic development which is insensitive to \hypers{}.
  We demonstrate that this benchmark is robust to statistical noise and obtains qualitatively similar results across repeated applications, even when using few samples.
  This robustness makes the benchmark computationally cheap to apply, allowing statistically sound insights at low cost.
  We demonstrate two example instantiations of the \shbabb{}, on a set of six small control environments (SC-\shbabb{}) and on the entire DM Control suite of 28 environments (DMC-\shbabb{}).
  Finally, to illustrate the applicability of the \shbabb{} to modern RL algorithms on challenging environments, we conduct a novel empirical study of an open question in the continuous control literature.
  We show, with high confidence, that there is no meaningful difference in performance between Ornstein-Uhlenbeck noise and uncorrelated Gaussian noise for exploration with the DDPG algorithm on the DMC-\shbabb{}.
\end{abstract}

\section{Introduction}
% --> Environment suites help to showcase generality of agents, but due to competitive benchmark pushing, we are now running experiments designed for best case performance
One of the major benefits of the Atari suite is the focus on more general reinforcement learning agents.
Numerous agents have been shown to exhibit learning across many games with a single architecture and a single set of \hypers{}, and to a lesser extent, OpenAI Gym \citep{brockman2016openai} and DM control suite \citep{tassa2018deepmind} are used in the same way.
As the ambitions of the community have grown, Atari and OpenAI Gym tasks have been combined into larger problem suites, with subsets of environments chosen to test algorithms.
In many ways we are back to where we started with Cartpole, Mountain Car and the like: where environment-specific \hyper{} tuning and problem subselection is prominent.
Instead of proposing a new and bigger challenge suite,
we explore a modification to standard empirical methodology for comparing agents across a given set of environments, complementing the existing empirical toolkit for investigating the scalability of deep RL algorithms.

% --> To get back to the generality purpose of domain suites, we need a benchmark designed for domain suites
In order to make progress towards impactful applications of reinforcement learning and the broader goals of AGI, we need benchmarks that clearly highlight the generality and reliability of learning algorithms.
Empirical work in Atari, Mujoco, and simulated 3D worlds typically use networks with millions of parameters, dozens of GPUs, and up to billions of samples \citep{beattie2016deepmind,espeholt2018impala}.
Many results are demonstrative, meaning that the primary interest is not the reliability and sensitivity, nor the resources required to achieve the result, rather that the result \emph{could} be achieved.
It is infeasible to combine these large scale demonstrations with \hyper{} studies and sound empirical methodology.
More evidence is emerging that such state-of-the-art systems
(1) rely on environment-specific design choices that are sensitive to minor changes to \hypers{} \citep{henderson2018deep,engstrom2019implementation}, (2) are less data efficient and stable compared with simple baselines \citep{vanhasselt2019when,taiga2019benchmarking}, and (3) cannot solve simple toy tasks without extensive re-engineering \citep{obando-ceron2021revisiting,patterson2021generalized}.
It is abundantly clear that modern RL methods can be adapted to a broader spectrum of challenging tasks---well beyond what was possible with linear methods and expert feature design.
However, we must now progress to the next phase of empirical deep RL research: focusing on generality and reliability.

% --> This better benchmark is part of a growing movement of improving empirical standards in RL
%Recently work has focused on raising
There is a growing movement to increase the standards of empirical work in RL.\@
Long before the advent of deep networks, researchers called out the environment overfitting that is rampant in RL and proposed sampling from parameterized variants of classic control domains to emphasize general methods \citep{whiteson2009generalized}.
Noisy results, inconsistent evaluation practices, and divergent code bases have fueled calls for more open-sourcing of agent architecture code, experiment checklists, and using more than three samples in our experiments \citep{henderson2018deep,pineau2020improving,patterson2023empirical}.
Recent work has highlighted our poor usage of basic statistics, including confidence intervals and hypothesis tests \citep{colas2018how,agarwal2021deep,patterson2023empirical}.
Finally, and most related to our work, \citet{jordan2020evaluating} proposed a methodology to better characterize the performance of an algorithm across environments, evaluated with randomly sampled \hypers{}.
We build on this direction, but focus on a simpler and more computationally frugal evaluation that examines the single best hyperparameter setting across environments, rather than a randomly sampled one, and allows for a smaller number of runs per environment.

% --> Table of badness
\begin{wrapfigure}[8]{l}{0.4\textwidth}
  \vspace{-0.1cm}
	\footnotesize
  % a little bit of magic to make wrapfigure think the label is a table, not a figure
  \makeatletter\def\@captype{table}\makeatother
  \caption{Chance of incorrect claims}\label{tab:failure}
  \centering
  \vspace{0.1cm}
  \begin{tabular}{lrrrr}
    & 3 runs & 10 & 30 & 100 \\
    \toprule
    Acrobot     & 47\% & 31\% & 22\% & 1\% \\
    Cartpole    &  7\% &  0\% &  0\% & 0\% \\
    CliffWorld  & 54\% & 19\% & 14\% & 0\% \\
    LunarLander & 16\% &  7\% &  1\% & 0\% \\
    MountainCar & 22\% &  9\% &  7\% & 0\% \\
    PuddleWorld & 18\% & 16\% &  8\% & 0\% \\
    \bottomrule
  \end{tabular}
\end{wrapfigure}

% --> Current empirical practices lead to a lot of mistakes, which make it difficult to build upon prior work
Experiments with many runs, \hypers{}, and environments can be computationally prohibitive, making these computational constraints a primary culprit for misleading or incorrect claims in RL experiments.
Typical strategies sacrifice one of these three axes to reduce costs, either using too few samples to draw statistically sound conclusions, providing an incomplete sensitivity analysis of the \hypers{}, or using a limited number of testbed environments to meaningfully evaluate the generality of the claims.
Table~\ref{tab:failure} illustrates the effect of using a limited number of seeds while tuning \hypers{}.
We ran four algorithms 250 times for every environment and \hyper{} setting in an extensive sweep to get a high confidence approximation of the correct ordering between algorithms.
We then used bootstrap sampling to simulate 10k papers---each using far fewer random seeds---and counted the frequency that incorrect algorithm orderings were reported.
Even with 30 runs in these small domains, incorrect rankings were \textbf{not} uncommon.
Further details are described in Section~\ref{sec:evaluation}.

It is surprising that 30 runs would be insufficient to reliably identify the correct ordering over four distinct algorithms for some environments.
This failure stems from a poor interaction between the statistical properties of RL algorithm performance and the challenge of identifying the best performing \hyper{} for an algorithm.
This is further exacerbated by modern RL algorithms, which require tuning an increasing number of \hypers{} and presenting increasingly complex \hyper{}-performance landscapes.
To combat this, several strategies have emerged in the literature including far more efficient tuning strategies than the commonplace gridsearch \citep{eggensperger2019pitfalls}, relying on default \hyper{} values \citep{schaul2016prioritized,wang2016dueling,vanhasselt2016deep,agarwal2021deep}, tuning \hypers{} on a subset of domains \citep{bellemare2013arcade}, or eroding standards of sufficient statistical power for publication \citep{henderson2018deep,colas2018how,agarwal2021deep}.

% --> Our goals are param sensitivity, stability over runs, cheap, easy-to-use, minimize advantage due to budget
In this paper, we evaluate the utility of selecting \hypers{} across environments using a methodology we call the \shb{} (\shbabb{}).
The basic idea is simple: we evaluate an algorithm across a set of environments using a single hyperparameter configuration in a two-stage approach.
Though conceptually simple, this methodology is not widely used.
We first address some nuances in the \shbabb{}, namely how to standardize performance across environments to allow for aggregation, how to allow for robust measures of performance, and finally how to reduce computation to make it more feasible to use the \shbabb{}.
We evaluate the effectiveness of the \shbabb{} itself by examining the stability of the conclusions from the \shbabb{} under different numbers of runs.
We then demonstrate that the \shbabb{} can result in different conclusions about algorithms compared to the conventional \emph{per-environment tuning} approach and the more recent approach of using a subset of environments for tuning.
Finally, we conclude with a larger demonstration of the \shbabb{} on the DM Control Suite.

\begin{figure}
  \includegraphics[width=\textwidth]{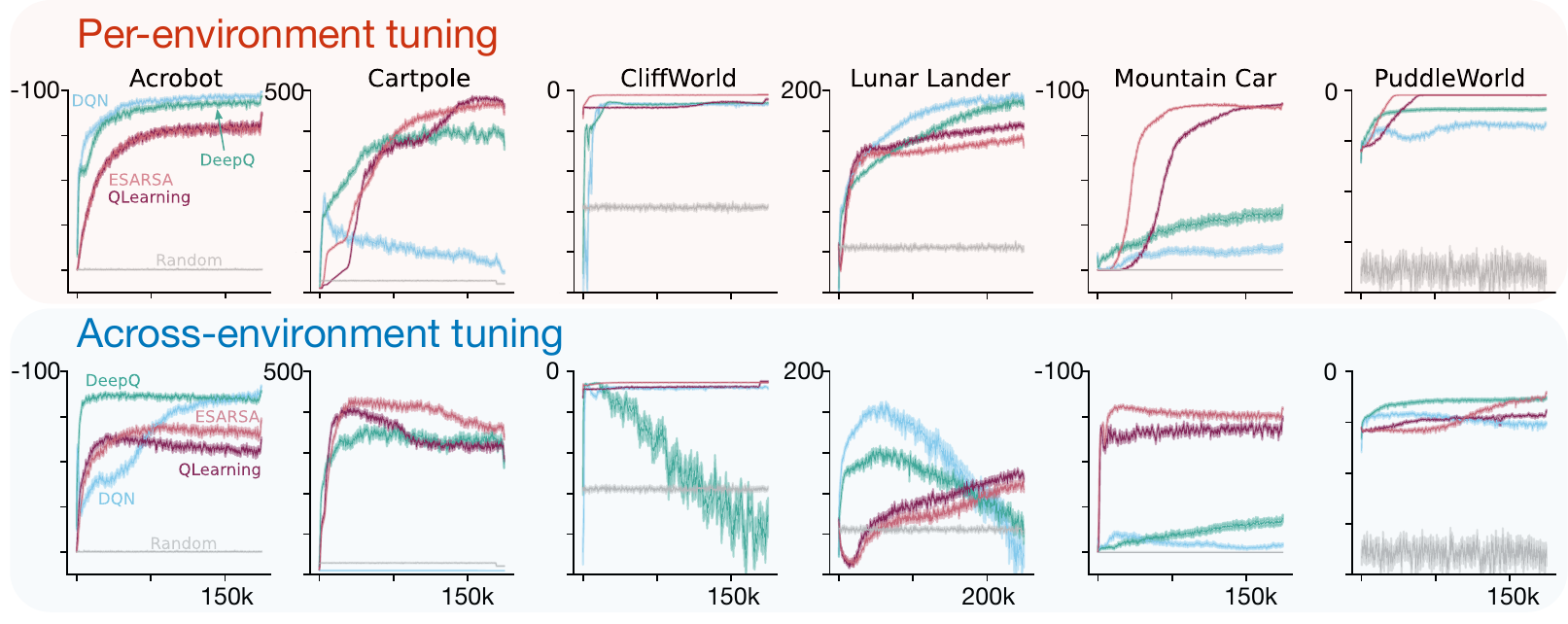}
  \caption{\label{fig:learning-curves}
    An example experiment comparing four algorithms across six different environments.
    Each learning curve shows the mean and 95\% confidence interval of 250 independent runs for each algorithm and environment.
    \Hypers{} are selected using three runs of every algorithm, environment, and \hyper{} setting.
    \textbf{Top} shows the learning curves when the best \hypers{} are chosen for each environment individually.
    \textbf{Bottom} shows the learning curves when \hypers{} are chosen according to the \shbabb{}.
  }
  \vspace{-0.4cm}
\end{figure}

\section{Contrasting Across-Environment versus Per-Environment Tuning}\label{sec:contrast}

In this section, we introduce the basic procedure for the \shbabb{} and provide an experiment showing how it can significantly change empirical outcomes compared to the conventional per-environment tuning approach. We provide specific details for each step later and here focus on outlining the basic idea and its utility.

\newcommand{\nruns}{n_{\text{tune}}}

The \shbabb{} consists of the following four steps summarized in the inset figure below. We assume we are given a set of environments and a set of \hypers{} for the algorithm we are evaluating.
\begin{wrapfigure}[10]{l}{0.5\textwidth}
\vspace{-0.1cm}
%\begin{figure}[htb!]
\centering
  \includegraphics[width=0.5\textwidth]{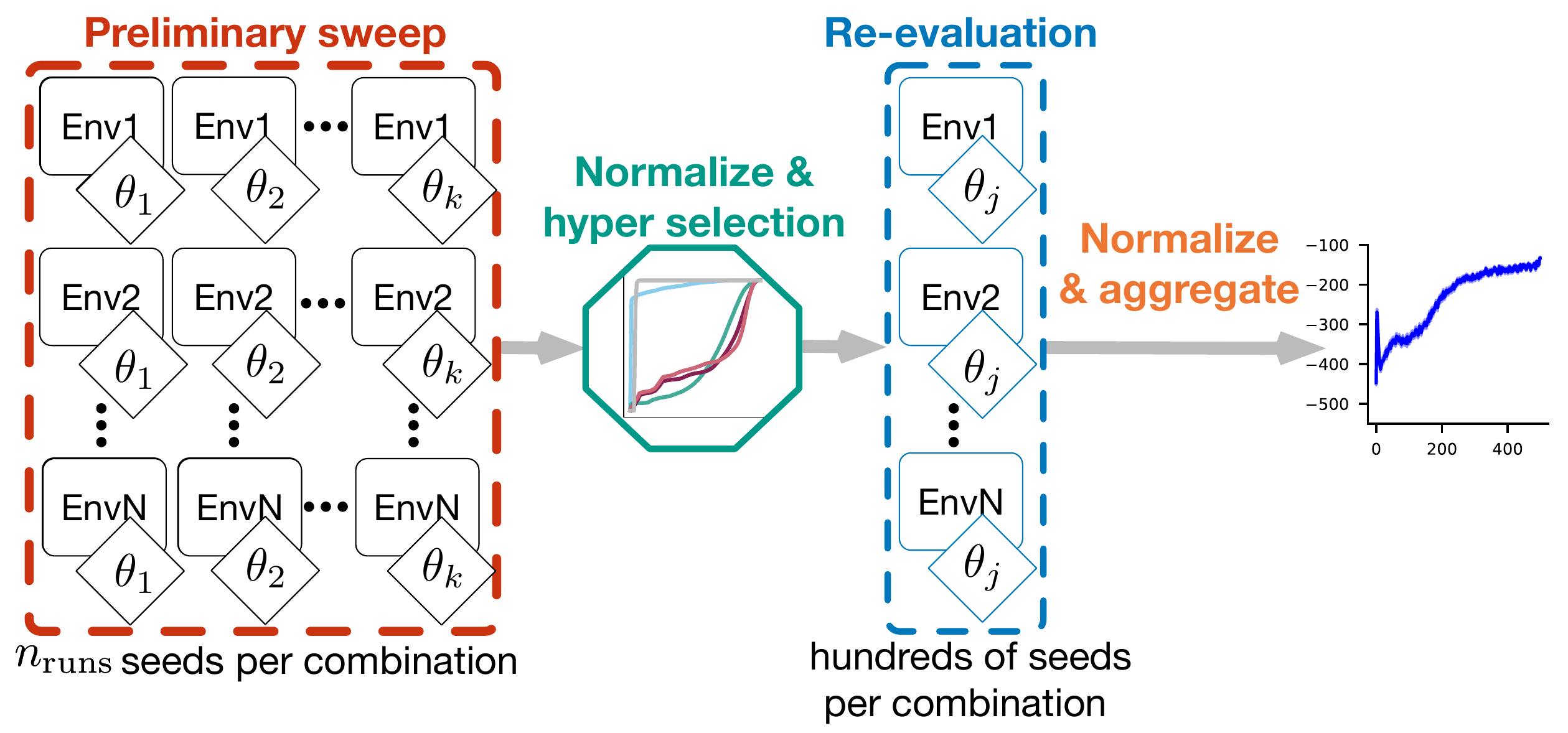}
  %\caption{
  %}
  \label{adam-fig}
\end{wrapfigure}
\textbf{Step 1 (Preliminary Sweep)} Run the algorithm for all \hypers{} and all environments, for $\nruns$ runs (i.e., $\nruns<30$) and record the performance of every combination. \\
\textbf{Step 2 (Normalization)} Normalize the scores across environments. We use CDF normalization, which is described in Section~\ref{sec:shb}.\\
\textbf{Step 3 (\Hyper{} Selection)} Select the \hyper{} setting with the highest score averaged across environments.  \\
\textbf{Step 4 (Re-evaluation)} With the single best \hyper{} setting, use many more runs in each environment (e.g. 100) to produce a more accurate estimate of performance.

% Get a more accurate estimate of performance of the single best \hyper{} in each environment using significantly more runs (e.g., 100).

The last step is more lightweight than it appears since only a single \hyper{} configuration is used for all environments.
By using a small $\nruns$ in the preliminary sweep, we save a significant amount of computational resources and can devote more resources to the re-evaluation step.
Detecting differences between hyperparameter configurations for each individual environment can be challenging, especially in the presence of noise.
For conventional \emph{per-environment tuning} to yield reliable and statistical sound results requires a large $\nruns$ for every algorithm, \hyper{}, and environment.
The \shbabb{}, by contrast, seeks to only detect differences in \hyper{} configurations \emph{across} environments, significantly reducing the necessary $\nruns$.

The benefit of combining normalized scores across environments is primarily statistical---averaging across more values typically results in a lower variance estimator.
However, it is well-established that finding a single \hyper{} configuration that works well across problems is challenging \citep{eggensperger2019pitfalls}.
This is precisely the goal of the \shbabb{}, to reduce the statistical hurdle of comparing algorithmic advances and focus on the challenge of designing algorithms which are less sensitive to their \hypers{}.

We illustrate this effect in Figure~\ref{fig:learning-curves}.
The per-environment tuning approach highlights the ideal behavior of an algorithm per environment, whereas the \shbabb{} highlights the (in)sensitivity of an algorithm across environments.
Experimental details can be found in Section~\ref{sec:evaluation}.
The environments are relatively simple (most coming from the classic control suite of OpenAI Gym \citep{brockman2016openai}) but difficult enough for our purposes: no one algorithm could reach near optimal performance in all environments.

In Figure~\ref{fig:learning-curves}, the \shbabb{} does not rank the algorithms differently than with per-environment tuning, but the \shbabb{} does alert us to potential catastrophic failure of some algorithms.
The neural network DeepQ agent performs terribly in Cliffworld and Lunar Lander under the \shbabb{}, but appears reliable under the per-environment approach. What is going on?
Forced to select only one \hyper{} across environments, the best outcome is to sacrifice performance in Cliffworld and Lunar Lander---achieving worse performance than a uniform random policy.

% --> We want to understand the joint distribution P(\Perf, \env, \param)
\section{Performance Distributions}\label{sec:perf-dists}
In this section, we describe the distribution and random variables underlying an RL experiment. This formalism allows us to reason about the summary statistics we consider for the \shbabb{} in the next section. We also visualize these distributions to provide intuition on the properties of the summary statistics of these distributions and the implications for the single performance numbers used in RL.

In an RL experiment, we seek to describe the performance distribution of an algorithm for each \hyper{} setting $\param \in \Params$, denoted as $\CP{\Perf, \env}{\param}$ where $\Perf$ is a random variable indicating the performance of an algorithm on a given environment, $\env \in \Envs$.
Most commonly, we report an estimate of the average performance conditioned on environment and \hyper{} setting, $\perf(\env, \param) \approxeq \CE{\Perf}{\env, \param}$ using a sample average and some measure of uncertainty about how accurately $\perf(\env, \param)$ approximates $\CE{\Perf}{\env, \param}$.

% --> We consider \env a random variable as well, here's what that lets us capture
The environment can be seen as a random variable for many RL experiments. The most common case is to specify a set of MDPs that the authors believe represent the important applications of their new algorithm. If results are uniformly aggregated across these environments, then this corresponds to assuming a uniform distribution over this set of environments. Other times, random subsets of environments from environment suites are chosen; the performance estimate on this subset provides an estimate of performance across the entire suite.
The idea of evaluating algorithms over a random sample of MDPs has been studied explicitly previously. For example, the parameters determining the physics of classical control domains were randomized and sampled to avoid domain overfitting \citep{whiteson2009generalized}, and randomly generated MDPs \citep{archibald1995generation} have been used to evaluate new algorithmic ideas \citep{seijen2014true,mahmood2014weighted,white2016investigating}. If we subselect after running the algorithms, then we bias the distribution over environments towards those with higher performance.

% --> Example distribution (Cartpole DQN cutoff=500 target_refresh=32 alpha=best)
\begin{wrapfigure}[10]{l}{0.25\textwidth}
  \vspace{-0.1cm}
  \centering
  \caption{\label{fig:perf-dist-demo}}
  \vspace{0.1cm}
  \includegraphics[width=0.25\textwidth]{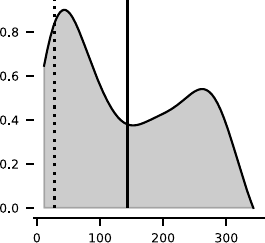}
  \vspace{-0.3cm}
\end{wrapfigure}
% --> Here's how we might generate a perf dist
Let us look at an example of these performance distributions to gain some intuition for estimating statistics like the expected performance.
Consider the action-value nonlinear control method DQN, using the Adam optimizer \citep{mnih2013playing,kingma2015adam}, on Cartpole \citep{barto1983neuronlike}.
We fix the \hyper{} setting $\param$ to the default values from \citet{raffin2019stable}. For this fixed environment, all randomness is due to sampling algorithm performance on this environment, namely sampling $G$ according to $\CP{\Perf}{\env, \param}$.
The performance, $G$, is the average episodic return over all episodes completed during 100k learning steps.
This environment is considered solved for $G > 300$.
We repeat this procedure for 250 independent trials to estimate the distribution $\CP{\Perf}{\env, \param}$, shown in Figure~\ref{fig:perf-dist-demo}, with x-axis possible outcomes of $G$ and y-axis the probability density.
The vertical solid line denotes mean performance, and the vertical dotted line denotes mean performance of a random policy.

% --> Here's what that perf dist says about stability over runs
Figure~\ref{fig:perf-dist-demo} is a typical example of the performance of an RL algorithm over multiple independent trials.
In this case, DQN is more likely to fail than to learn a policy which solves this relatively simple environment.
It is common practice to run an RL algorithm for some number of random seeds---effectively drawing samples of performance from this distribution---then reporting the mean over those samples (solid vertical line).

There are two implications from observing this bimodal performance distribution. First, using the expected value of this distribution as the summary statistic does not aptly demonstrate that the poor performance of DQN on Cartpole is due to occasional catastrophic failure---performing worse than or equivalent to a random policy.
Instead, mean performance might lead us to wrongly conclude that DQN on Cartpole usually finds a sub-optimal, yet better than random, policy. An alternative might be to consider percentile statistics or, if the goal is to evaluate mean performance, to avoid drawing strong conclusions about individual runs.

% --> These non-normal distributions take a lot of runs to appropriately estimate
If the goal is to report mean performance, then a second issue arises.
Estimating the mean of these non-normal performance distributions can be challenging.
In Figure~\ref{fig:perf-dist-demo}, approximately 70\% of the density is around a mode centered at 20 return, and the remaining 30\% is around a mode centered at 250 return.
As a result, sample means constructed with only three runs are varied and skewed.

% --> Reporting highest average performance presents yet another challenge.
Further, to report the average performance of the best performing \hyper{}---that is $\max_{\param \in \Params} \CE{\Perf}{\param, \env}$---we must first reliably estimate the conditional expected performance for each \hyper{}.
Computing this expectation can require a large number of samples to obtain a reasonable estimate for each \hyper{}.
This results in a tradeoff between measuring sensitivity and reliability: between the breadth of \hyper{} settings that can be studied and the accuracy to with which we can feasibly evaluate each \hyper{}.

% --> Performance distributions for a few different stepsizes, all other parameters fixed
\begin{wrapfigure}[10]{l}{0.28\textwidth}
  %\vspace{-0.4cm}
  \centering
  \includegraphics[width=0.28\textwidth]{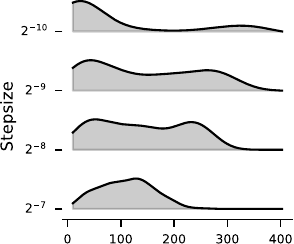}
  %\vspace{0.3cm}
  %\caption{\label{fig:alpha-dist-demo} Cartpole}
\end{wrapfigure}
% --> These weird distributions make it hard to pick \hypers{}
The summary statistic used to select \hypers{} also interacts with the form of the performance distribution.
In the inset figure on the {\bf left} we show the performance distribution across four different choices of stepsize parameter of DQN in {\bf Cartpole}. %the subplot corresponding to $\text{stepsize}=2^{-9}$ is same distribution as in Figure~\ref{fig:perf-dist-demo}.
If we are interested only in the highest best case performance, then $2^{-10}$ is preferred.
However, if we are particularly concerned with reducing the chances of catastrophic failure (i.e., highest worst case performance), then a stepsize $2^{-7}$ is preferred.
The most common case is to report results for the stepsize with the highest average performance.
In this case, a stepsize of $2^{-9}$ would be preferred.

% --> Performance distributions for a few different stepsizes on puddleworld
% --> What happens if we look at another domain?
\begin{wrapfigure}[10]{R}{.28\textwidth}
%   \vspace{-0.4cm}
  \centering
  \includegraphics[width=.28\textwidth]{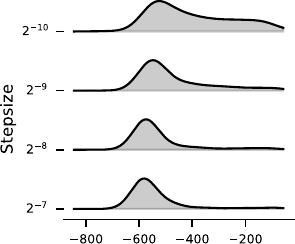}
  %\vspace{0.3cm}
  %\caption{\label{fig:alpha-dist-demo-2} Puddle World}
\end{wrapfigure}
These performance distributions can also look quite different for different environments, even with the same algorithm.
For Cartpole (above), the distribution is increasingly long-tailed with smaller stepsizes.
For {\bf Puddle World}, shown in the inset figure on the {\bf right}, the distributions are always bimodal with one mode around -600 return and a second mode around -200 return.
With smaller stepsizes, the density around the better performance mode increases, shifting the mean of the distribution. Peak performance does not change; rather the probability that DQN has a good run is higher with small stepsizes. This  analysis of performance distributions raises an important question:
do current RL algorithms have consistent \hyper{} settings which perform well across many environments?

% --> Why is this a problem?
%As we increase the number of environments we consider, we notice that there are few consistent patterns in the performance distributions across stepsizes for DQN.\@
%These results can be found in Appendix~\ref{app:additional-results}.
%In some environments, like Acrobot, the stepsize with best average performance is much larger than for either Cartpole or Puddle World above.
%In other environments, like Lunar Lander, the performance distributions appear approximately normal for all choices of stepsize making the selection process very easy on this environment.

\section{The \shb}\label{sec:shb}

% --> In this section we describe a new benchmark for evaluation of RL algorithmic design across environments
In this section, we describe the \shb{} (\shbabb{}) in detail.
Although it seems natural to evaluate across environments, standard empirical practice in RL is not done this way.
Understanding across-environment sensitivity aligns nicely with the intent of sensitivity analysis: elucidating how well an algorithm might perform on new environments without extensive \hyper{} tuning.
We argue that the \shbabb{} 1) better aligns empirical practice with the goals of applied RL, 2) is computationally feasible even in complex environments, 3) provides novel insights on old ideas (even with small environments), and 4) reduces the chances of accidentally publishing incorrect conclusions due to statistical noise.

% --> High-level \shb{} design
The first step of the \shbabb{} ({\bf preliminary sweep}) is to draw a small number of samples $\nruns$ from $\CP{\Perf}{\param, \env}$ for every \hyper{} setting and environment and get the summary estimate $\perf(\env, \param)$ from those samples. Typically, we compute $\perf(\env, \param)$ as a sample average to estimate $\CE{\normalizeE{\Perf}}{\env,\param}$, where $N_\env : \Re \to \Re$ is a \textbf{normalization} function that we describe below. Then we aggregate across environments to estimate $g(\theta) \approx \E{\CE{\normalizeE{\Perf}}{\env,\param}}$, where the outer expectation is with respect to environments. Then we {\bf select} a single \hyper{} setting with $\shbparam = \argmax_{\param \in \Params} g(\theta)$.
Finally, we draw a large number of samples from $\CP{\Perf}{\shbparam, \env}$ for every environment and report the same summary statistics $\perf(\env, \shbparam)$ and $\perf(\shbparam)$ ({\bf re-evaluation}).
%For example, if we use mean for performance in the environment and across environments, we estimate $\E{\CE{\normalize{\Perf}}{\env,\shbparam}}$, now with a large number of samples to get a much more accurate estimate.

% --> Scaling performance
Generally, we cannot expect each environment to produce normalized performance numbers, so to compute the expectation across environments we must first normalize the performance measures.
A comprehensive discussion of normalization methods is given in \citet{jordan2020evaluating}.
We use a lightly modified version of the CDF normalization method from \citet{jordan2020evaluating}, $\normalizeE{\Perf} = \text{CDF}(\Perf, \env)$, which is highly related to probabilistic performance profiles \citep{barreto2010probabilistic}.

To compute the CDF normalization, we first collect the performance $g$ of each algorithm and \hyper{} into a pool $\mathcal{P}_{\env}$ for each environment $\env$.
Then given some arbitrary score $x$ from environment $\env$, the CDF normalization is
\begin{equation*}
\text{CDF}(x, \env) = \frac{1}{|\mathcal{P}_{\env}|} \sum_{\perf \in \mathcal{P}_{\env}} \textbf{1}(g < x)
\end{equation*}
where $\textbf{1}$ is the indicator function. This mapping says: what percentage of performance values, across all runs for all algorithms and all \hyper{} settings, is lower than my performance $x$ on this particular environment $\env$? For example, if $\text{CDF}(x, \env) = 0.25$, then this agent's performance is quite low in this environment, as only 25\% of other agents' performance was worse across agents tested. This normalization accounts for the difficulty of the problem, and reflects relative performance amongst agents tested.
Note that this normalization uses an empirical CDF, rather than the true CDF for the environment and set of hyperparameters and agents. This means there is a small amount of bias when estimating $\E{\CE{\normalizeE{\Perf}}{\env, \param}}$. This bias dissipates with an increasing numbers of samples and equally impacts all compared algorithms.

% --> cheaper than per-environment maxing when done correctly
Selecting \hypers{} with the \shbabb{} can require significantly fewer samples compared with conventional per-environment tuning.
Per-environment tuning requires a sufficiently accurate estimate of the conditional expectation $\CE{\Perf}{\env, \param}$ for every $\param \in \Params$ and for every $\env \in \Envs$, requiring a number of runs proportional to $|\Params||\Envs|$.
The \shbabb{}, on the other hand, requires only an accurate estimate of $\CE{\normalizeE{\Perf}}{\param} = \E{\CE{\normalizeE{\Perf}}{\env, \param}}$ which requires a number of runs proportional only to $|\Envs|$. By designing a process which selects \hypers{} first using a smaller number of runs, we can reserve more computational resources for re-evaluation. Once we select the best \hypers{}, the cost of collecting samples is independent of $\Theta$, and so we can decouple the precision of our performance estimate from the number of \hyper{} settings that we evaluate for each algorithm.

%% --> Other summarizations of environments
% MARTHAC: Removed for now for space. Sort of obvious from above.
%It is possible to modify the \shbabb{} using different performance statistics across environments or \hypers{}.
%For instance, in robust RL it is common to seek an algorithm which performs well on a worst case environment \citep{delage2007percentile}.
%In our context, this can be achieved by replacing the expectation over environments with a minimization or a percentile statistic.
%The remainder of the procedure would remain the same and would answer questions of the form: ``Which algorithm has a single \hyper{} setting which performs well on the worst-case environment within this set of environments?''

Finally, we can contrast this benchmark with a recent evaluation scheme that uses random hyperparameter selection \citep{jordan2020evaluating}. In order to capture variation in performance due to \hyper{} sensitivity, \citet{jordan2020evaluating} treats \hypers{} as random variables and samples according to an experimenter-designated distribution over \hypers{}, reporting the mean and uncertainty with respect to this added variance, similar to the procedure used in \citet{jaderberg2016reinforcement}.
This evaluation methodology provides some insight into the difficulty of tuning, though requires a sensible distribution over \hypers{} to be chosen. The \shbabb{}, on the other hand, asks: is there a \hyper{} setting for which this algorithm can perform well across environments? It motivates instead identifying that single \hyper{}, and potentially fixing it in the algorithm, or suggesting that the algorithm needs to be improved so that such a \hyper{} could feasibly be found. Both of these strategies help identify algorithms that are difficult to tune, but the \shbabb{} is easier to use and computationally cheaper.

\section{Evaluating the \shb{}}\label{sec:evaluation}
% --> In this section we evaluate the evaluation
In this section, we evaluate the \shbabb{} by comparing four algorithms across several classic control environments.
Evaluating the reliability of an evaluation strategy is challenging as we need to understand the probability that the \shbabb{} leads to drawing incorrect conclusions---similar to ensuring that a methodology for computing 95\% confidence intervals do in fact capture the mean value 95\% of the time.
To achieve this, we gather an extensive dataset of 250 samples for every algorithm, \hyper{}, and environment. We treat this large dataset as the source-of-truth and draw (with replacement) subsamples of this dataset to simulate a single application of the \shbabb{}. Each of these simulated applications of the \shbabb{} can be thought of as a single paper using the \shbabb{} to compare multiple algorithms.
We then estimate what proportion of those papers using the \shbabb{} identify the correct ordering of algorithms.
Nominally, using 95\% confidence intervals, we would expect to identify the correct ordering 95\% of the time.

For this simulation of the \shbabb{}, we use a dense and exhaustive gridsearch over \hypers{}.
By using an exhaustive gridsearch, we can ensure that a high-performing \hyper{} configuration is captured in the set of tested \hypers{}---though at high computational cost.
Using a gridsearch also greatly simplifies the statistical simulation strategy used to evaluate the \shbabb{} over many simulated papers, without changing conclusions about the \shbabb{} itself.
However, the \shbabb{} is agnostic to the \hyper{} configuration strategy and typically a gridsearch is not the most computationally efficient approach.

For this evaluation, we require environments where hundreds of independent samples of performance can be drawn across a large \hyper{} sweep in a computationally tractable way.
We emphasize that this is not a general requirement of the \shbabb{} and is required only in this case of evaluating the \shbabb{}'s responsiveness to perturbations in the experimental process.
Because these classic control environments are cheap to run and provide meaningful insights in differentiating modern RL algorithms \citep{obando-ceron2021revisiting}, we name this specific benchmark the Small Control \shbabb{} (SC-\shbabb{}).
In Section~\ref{sec:big-demo} we provide a realistic demonstration of the \shbabb{} on a larger dataset with a more complex algorithm.

% --> Describe algorithms
\textbf{Algorithms.}
For the following investigations, we compare two deep RL algorithms based on DQN \citep{mnih2013playing} and two control algorithms based on linear function approximation using tile-coded features \citep{sutton2018reinforcement}.
The deep RL algorithms, DQN and DeepQ, differ only in their loss: DQN uses a clipped loss and DeepQ uses a mean squared error.
%function that they each optimize.
%DQN uses a clipped loss, similar to a Huber loss \citep{mnih2013playing,huber1992robust} and DeepQ uses a mean squared error.
For the two tile-coding agents, QLearning is off-policy and bootstraps using the greedy action, while ESARSA is on-policy and bootstraps using an expectation over actions.
Further details on the algorithms can be found in Appendix~\ref{app:experiment-details}.

% --> Describe environments
\textbf{Environments.}
The SC-\shbabb{} consists of a suite of classic control environments commonly used in RL: Acrobot \citep{sutton1996generalization}, Cartpole \citep{barto1983neuronlike,brockman2016openai}, Cliff World \citep{sutton2018reinforcement}, Lunar Lander \citep{brockman2016openai}, Mountain Car \citep{moore1990efficient,sutton1996generalization}, and Puddle World \citep{sutton1996generalization}.
We used a discount factor of $\gamma=0.99$ and a maximum episode length of 500 steps (except in Cliff World which had a maximum length of 50 steps).
We ran all algorithms for 200k learning steps on each environment except Lunar Lander, where we used 250k learning steps to ensure all algorithms have reliably converged.
Further details motivating this choice of environments can be found in Appendix~\ref{app:small-control-environments}.

% --> Describe meta-params

\textbf{\Hypers{}.}
For all algorithms we swept eight stepsize values, $\alpha \in \{2^{-12}, 2^{-11}, \ldots, 2^{-5}\}$ for the deep RL algorithms and $\alpha \in \{2^{-9}, 2^{-8}, \ldots, 2^{-2} \}$ for the tile-coded algorithms.
The deep RL algorithms used experience replay and target networks, so we swept over replay buffer sizes of $\{2000, 4000\}$ and target network refresh rates of $\{1, 8, 32\}$ steps where a one step refresh indicates target networks are not used.
The algorithms with tile-coding learn online from the most recent sample; we select number of tiles in each tiling in $\{2, 4, 8\}$ and number of tilings in $\{8, 16, 32\}$.
%The tile-coded algorithms were learned online from the most recent sample and without target networks, instead we swept over parameters of the tile-coder.
%We swept over the number of tiles in each tiling in $\{2, 4, 8\}$ and number of tilings in $\{8, 16, 32\}$ for all environments.
More details on the other \hypers{} and design decisions are in Appendix~\ref{app:experiment-details}.

\begin{figure}[t]
  \centering
  \includegraphics[width=0.8\textwidth]{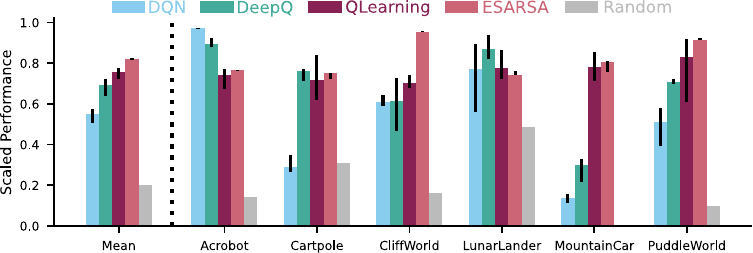}
%  \vspace{0.2cm}
  \caption{\label{fig:bootstrap-shb}
    Applying the \shbabb{} to 10k simulated experiments.
    Error bars show 95\% bootstrap confidence intervals.
    Although only three runs were used to select \hypers{}, conclusions about algorithm ranking using the \shbabb{} are perfectly consistent across all 10k experiments.
  }
  \vspace{-0.1cm}
\end{figure}

\textbf{Variance over simulated experiments.}
We start by demonstrating the low variance of conclusions over 10k simulated applications of the \shbabb{}.
We simulate applying the \shbabb{} with three random seeds for every algorithm, environment, and \hyper{} to first select \hypers{}, then using 250 random seeds to evaluate the performance of each algorithm on each environment with the selected \hyper{} configuration.
We compare the outcomes of each application of the \shbabb{} to estimate the variance in conclusions in Figure~\ref{fig:bootstrap-shb}.

Using the full dataset, the true ordering of algorithms from best to worst is ESARSA, QLearning, DeepQ, and DQN.
Every simulated application of the \shbabb{} detected this ranking successfully.
Conclusions on individual environments are less consistent, though this is to be expected.
The \shbabb{} sacrifices the ability to draw conclusions about the ordering of algorithms on individual domains by setting $\nruns$, the number of tuning seeds, to be very small.
In reality, the use of small $\nruns$ to draw conclusions on individual domains is common practice and Figure~\ref{fig:bootstrap-shb} well demonstrates how this practice can be misleading.

We provide more insight into the difficulty of selecting a single hyperparameter across problems, in Appendix \ref{app_hypersensitivity}. We additionally show that the distribution of selected \hypers{} with the \shbabb{} is narrow and consistent over simulated experiments, unlike parameters chosen independently for each environment.
Because conclusions are often drawn by aggregating results over environments---either formally as in the \shbabb{} or informally by counting the number of environments where an algorithm outperforms others---reporting results over a consistent and narrow distribution of \hypers{} leads towards lower variance claims and greater reproducibility. We include results selecting hyperparameters according to the worst-case performance across environments in Appendix \ref{app_worstcase}; the results are highly similar, albeit slightly lower variance.

The cost of running a single experiment represented in Figure~\ref{fig:bootstrap-shb} is quite low.
The deep RL algorithms test 48 \hyper{} settings at a cost of 20 minutes per run, while the tile-coded algorithms test 72 settings at the cost of two minutes per run.
Timings are with respect to an older 2.1Ghz Intel Xeon processor.
This comes out to a total of 1762 hours of CPU time to complete three runs for \hyper{} selection and 250 runs for evaluation, cheaper than the experiment using 10 runs and conventional per-environment tuning shown in Table~\ref{tab:failure} which cost approximately 2208 hours.
The \shbabb{} successfully detected the correct ordering of algorithms in every trial, while the conventional per-environment tuning experiment failed to detect the correct ordering with surprising frequency.

\textbf{The \shbabb{} is a less optimistic measure of performance.}
A motivating factor for the \shbabb{} is providing a more challenging benchmark to test across-environment insensitivity to selection of \hypers{}.
Because algorithms are limited to selecting a single champion \hyper{} setting---as opposed to selecting a new \hyper{} setting for every environment---we expect a considerable drop in performance under the \shbabb{}.
We evaluate the extent of this performance drop for our four algorithms by first computing near optimal parameters $\param^* \in \Params$ for each environment using the full 250 random seeds to obtain high confidence estimates of average performance $\CE{\normalizeE{\Perf}}{\env, \param^*}$.
We then apply the \shbabb{} to select \hypers{} for each algorithm using three random seeds for 10k simulated experiments.
We report sample estimates of $\CE{\normalizeE{\Perf}}{\env, \param^*} - \CE{\normalizeE{\Perf}}{\env, \shbparam}$.

\begin{figure}[t]
  \centering

  \includegraphics[width=0.8\textwidth]{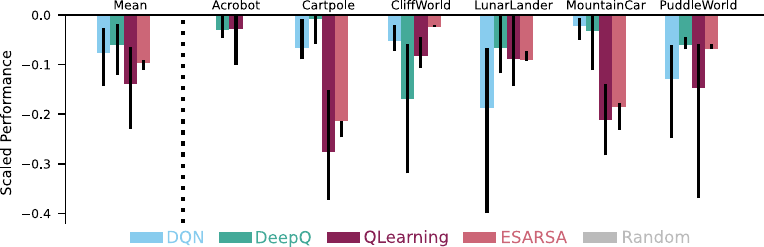}
  \vspace{0.01cm}
  \caption{\label{fig:perf-drop-shb}
    The change in performance for each algorithm on every environment when using the \shbabb{} versus conventional per-environment tuning.
    A larger drop in performance indicates a larger degree of environment overfitting when results are reported with per-environment tuning.
    Error bars show 95\% confidence intervals over 10k bootstrap samples.
   % The mean column shows the average over differences on individual domains.
  }
\end{figure}

In Figure~\ref{fig:perf-drop-shb} we can see there is substantial drop in reported performance when using the \shbabb{} versus per-environment tuning. The variance is high, indicating that for some runs, the performance drop was substantial: almost 0.4 under our normalization between [0,1].
Algorithms with a large drop in performance indicate more environment-specific overfitting under per-environment tuning.
Because we swept over many more \hyper{} settings for the tile-coding algorithms than for the deep RL algorithms---72 settings versus 48 settings---it is unsurprising that per-environment tuning led to far more environment overfitting in the tile-coding algorithms.

%Figure~\ref{fig:perf-drop-shb} calls to question the common practice of ablating novel \hypers{} when introducing a new algorithmic development.
%Ablating more \hyper{} settings for new algorithms, but using default parameters or a smaller parameter study for benchmark algorithms, can lead to considerable over-claims in performance simply through \hyper{} tuning.
%As seen in Figure~\ref{fig:perf-drop-shb}, the \shbabb{} helps defend against this form of overfitting by significantly under-reporting performance of algorithms that used more extensive \hyper{} tuning.
%While it is still possible to use extensive \hyper{} tuning to obtain misleading results with the \shbabb{}, it becomes computationally expensive to the point of being impractical.

\textbf{Tuning on a subset of environments.}
An empirical practice that is highly related to the \shbabb{} is using a subset of environments to select \hypers{}, then reporting the performance of the selected \hypers{} across an entire suite of environments. We refer to this practice as \textit{subset-\shbabb{}}.
This practice is used in the Atari suite for example, where it was suggested to use five of the 57 games for \hyper{} tuning \citep{bellemare2013arcade}.
To investigate the variance of conclusions using the subset-\shbabb{}, we run 10k simulated experiments using two of our six environments to select \hypers{}.
For each of the simulated experiments, we randomly select two environments to use for \hyper{} selection.
To reduce the variance, we allow each algorithm 100 runs of every \hyper{} setting on every environment to perform \hyper{} selection, then evaluate the performance on the full 250 runs for the \hyper{} selected by the subset-\shbabb{}.
More results, including with varying number of runs and environments used for \hyper{} selection, can be found in Appendix~\ref{app:additional-results}.

\begin{figure}[t]
  \centering

  \includegraphics[width=0.8\textwidth]{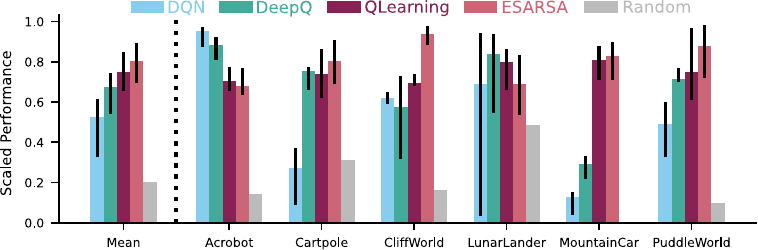}
  \vspace{0.02cm}
  \caption{\label{fig:oos-shb}
    Performance of each algorithm over 10k bootstrap samples, where sample means are computed with 100 runs.
    Each bootstrap sample randomly selects two environments for \hyper{} tuning, then evaluates the chosen \hyper{} setting on all six environments with 250 runs.
    Error bars show 95\% bootstrap confidence intervals.
  }
\end{figure}

In Figure~\ref{fig:oos-shb}, we see that the ordering of algorithms is extremely high-variance---especially compared to Figure~\ref{fig:bootstrap-shb} which uses all six environments to select \hypers{} and only three runs.
This result also illustrates large differences between individual environments, where the variance on Lunar Lander---especially for DQN---suggests that \hypers{} selected for other environments are likely to cause worse-than-random performance on Lunar Lander.
At least among the four tested algorithms, it is clear that \hyper{} sensitivity is too high to use environment subselection to reduce the computational burden of \hyper{} tuning.

\textbf{Bias of the \shbabb{}.}
Both the \shbabb{} and conventional per-environment tuning use biased sample estimates due to the maximization over \hypers{}.
The bias due to maximization over random samples is exaggerated both as the set $\Params$ grows and as the number of samples used to evaluate $\CE{\Perf}{\env, \param}$ shrinks.
\begin{wrapfigure}[15]{l}{0.33\textwidth}
  \vspace{-0.3cm}
  \centering
  \includegraphics[width=0.33\textwidth]{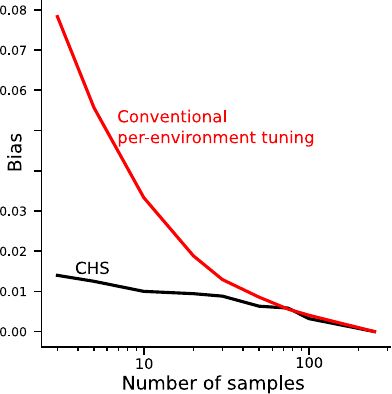}
  \vspace{-0.2cm}
  \caption{\label{fig:consistency} Bias of the \shbabb{} vs. per-environment tuning.}
\end{wrapfigure}
We first estimate the true per-environment maximizing parameters $\param^*$ and the true \shbabb{} parameter $\shbparam^*$ using 250 samples for every \hyper{} setting and environment.
We then resample three samples per \hyper{} and environment to simulate an experiment using three seeds to compute sample averages, we select the maximizing parameter of these sample averages, $\hat{\param}$, and we report $\CE{\Perf}{\env, \param^*} - \CE{\Perf}{\env, \hat{\param}}$.
The corresponding procedure is used for the \shbabb{}.

In Figure~\ref{fig:consistency}, we report the bias of each procedure applied to DQN and the small control domain suite.
On the vertical axis we report the bias and on the horizontal axis we show the number of random seeds used to select \hypers{}.
As both procedures approach a sufficiently large number of samples to select \hypers{}, the bias of these procedures approaches zero.
However when using few random seeds---for instance ten or fewer as is common in the literature---the bias of the conventional method is several times larger than that of the \shbabb{}.
As a result of this overestimation bias, it is common for results in the literature to present highly optimistic results especially for algorithms with more \hypers{}.

\section{A Demonstrative Example of Using the \shbabb{}}\label{sec:big-demo}
We finish with a large-scale demonstration of our benchmark across the 28 environments of the DMControl suite \citep{tassa2018deepmind}, which we will call the DMC-\shbabb{}.
For this comparison, we test an open hypothesis in the continuous control literature: does Ornstein-Uhlenbeck (OU) noise \citep{uhlenbeck1930theory} improve exploration over naive uncorrelated Gaussian noise?
Autocorrelated noise for exploration was shown to be beneficial for robotics \citep{wawrzynski2015control}, inspiring the use of an OU noise process for DDPG \citep{lillicrap2016continuous}, where a single set of \hypers{} was used across 20 Mujoco environments using five seeds.
Later work replaced OU noise with Gaussian noise, noting no difference in performance \citep{fujimoto2018addressing,barth-maron2018distributed}, but without empirical support for the claim.
To the best of our knowledge, no careful empirical investigation of this hypothesis has yet been published.

\begin{figure}
  \centering
  \includegraphics[width=.75\textwidth]{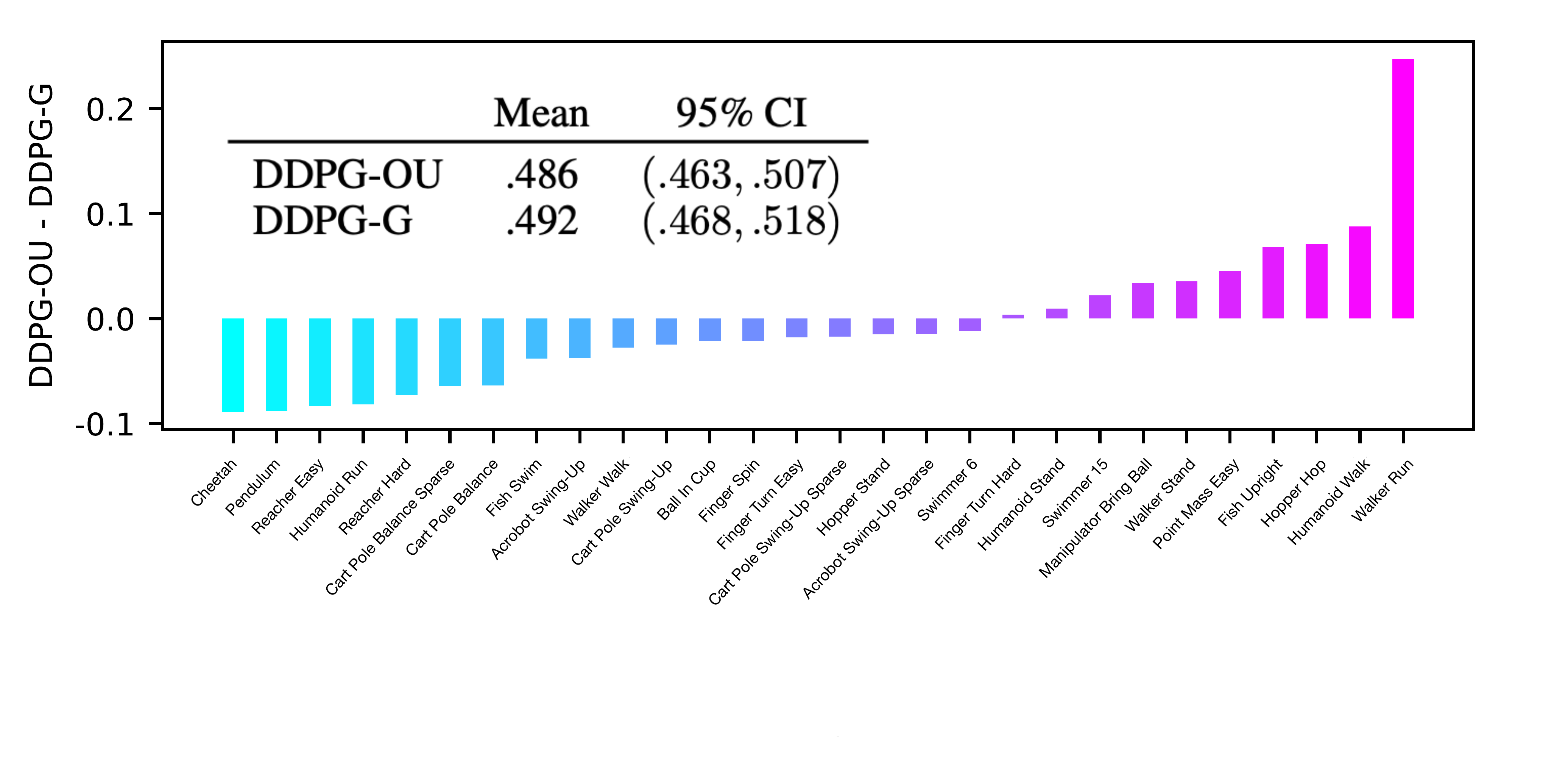}
  \vspace{-0.8cm}
  \caption{\label{fig:ddpg-ou-envs}
    Comparing DDPG using OU noise vs. Gaussian noise across the DMControl suite. The inset table shows the mean performance with 95\% confidence interval for the two versions of DDPG used in these experiments. Visualized in the bar plot is the performance of DDPG with OU noise, per environment in the suite, considering DDPG with Gaussian noise as a baseline.
    }
\end{figure}

To apply the DMC-\shbabb{}, we first evaluate 36 \hyper{} settings with three runs per environment, for a total of 84 runs to estimate $\CE{\normalizeE{\Perf}}{\param}$ for each $\param \in \Params$.
Then we use 30 runs to evaluate the chosen $\shbparam$ for a total of 840 runs to estimate $\CE{\normalizeE{\Perf}}{\shbparam}$.
We report the swept \hypers{} as well as the selected $\shbparam$ in Appendix~\ref{app:dmcontrol-results}.
We use 1k bootstrap samples to compute confidence intervals and report the overall findings in %Table~\ref{tab:ddpg-ou-g}.
the table in Figure~\ref{fig:ddpg-ou-envs}.
We find that OU noise does not outperform Gaussian noise on the DMC-\shbabb{}.
Considering even the extremes of the confidence intervals there is no meaningful difference in performance between these exploration methods, suggesting further runs would be unlikely to change our conclusion.
We visualize the performance of OU noise on the complete suite, considering Gaussian noise experiments as a baseline in Figure~\ref{fig:ddpg-ou-envs}. This visualization summarizes whether, and to what degree, OU noise improves upon Gaussian noise in each environment of the DMControl suite. In only 10 of the 28 environments, OU noise improves upon Gaussian noise, with a large improvement only in the \textit{WalkerRun} environment.
Additional results are included in Appendix~\ref{app:dmcontrol-results}.

\section{Conclusion}
In this work, we introduced a new benchmark for evaluating RL algorithms across environments, but perhaps more important are the insights we gained.
Of the five algorithms we tested (including DQN and DDPG), none exhibited good performance on our \shbabb{} benchmark; aligning with the common view that we do not yet have generally applicable RL algorithms.
The \shbabb{} benchmark produces reliable conclusions with only three runs in the preliminary sweep while providing a new challenging aspect to small computationally-cheap environments, allowing small university labs and tech giants alike to conduct rigorous and meaningful comparisons.
Finally, prior work has disagreed on the benefit of using OU or Gaussian noise in DDPG on Mujoco-based environments.
Perhaps some combination of too few runs, using default \hypers{}, or problematic environment sub-selection yielded conflicting results.
Our results with \shbabb{} suggest there is no significant performance difference across a suite of 28 Mujoco environments, putting this debate to bed.
The \shbabb{} benchmark can play a role uncovering falsehoods and resolving disputes.

The \shbabb{} is a general procedure for evaluating performance across environments.
We provide two example instantiations of the \shbabb{}, the SC-\shbabb{} for discrete action control on small domains and the DMC-\shbabb{} for continuous control on large simulated environments, however the \shbabb{} can also be extended to use arbitrary environment sets to allow targeted evaluation across environments with certain desireable properties.
For example, the taxonomies of Atari games identified in \citet{bellemare2016unifying}, the off-policy evaluation environments used in \citet{sutton2009fast}, or the taxonomy of exploration environments from \citet{yasui2019empirical} are each sets of environments that have been previously identified and used across the literature.
Applying the \shbabb{} to any one of the environment sets provides a new challenge, and in some small way can push us towards generally applicable RL agents.

%In this work, we introduced a new benchmark for encouraging development of RL algorithms which are insensitive to \hypers{} across environments.
%The \shbabb{} represents a challenge for algorithm development in RL, to design algorithms which perform well across many environments without extensive \hyper{} tuning.
%In order to scale alongside increasingly large benchmark environments, the \shbabb{} is computationally efficient, allowing statistically sound insights at a small cost.
%Finally, we demonstrate the applicability of the \shbabb{} on a large and open-question, providing novel insights on the DMControl suite.

\clearpage
\pagebreak

\bibliographystyle{mybibstyle}
\bibliography{paper}

\clearpage
\pagebreak
\appendix

\section{Ethical considerations}\label{app:ethics}
Because the \shb{} is applied to experimenter-picked domains, it inherits the biases and ethical considerations of the studied domains.
However, the primary goal of the \shbabb{} is to reduce algorithm design decisions which lead to overfitting to specific attributes of studied domains.
An example could be the use of layer-normalization within neural networks, which highly disproportionately favor pixel-based domains.
If the \shbabb{} utilized only pixel-based domains, the results would still follow the bias of the experiment designer; however if even a single non-pixel domain was included in the benchmark, the biased design decision to use layer-normalization would negatively impact the outcome of this particular algorithm on this benchmark.
Previous empirical practices are more likely to permit this form of biased design due to statistical noise or domain subselection.

A major motivation for this paper is to advocate that meaningful and sound experiments in RL are achievable at an inclusive and low cost.
Running a complete experiment with four algorithms on the six small control environments used in Section~\ref{sec:evaluation} would cost approximately \$40 USD at current AWS EC2 prices and would complete in approximate two days.
This experiment is complete, sound, and provides a meaningful ranking of four comparison algorithms; even detecting performance differences across minute algorithmic differences.
However, the \shbabb{} is not perfectly resilient to gaming with extensive \hyper{} tuning.
Consider the highly common scenario where we wish to advocate for one algorithm over competitive baselines, then performing extensive tuning or multiple iterations of tuning with the \shbabb{} still gives advantage to labs with greater access to resources.

For the studies performed in Section~\ref{sec:evaluation}, we required far more compute than would be typical of a study utilizing the \shbabb{}.
To evaluate the effectiveness of the \shbabb{} required sufficient data to simulate repeated applications of \shbabb{} on new data.
Our results were obtained using a cloud CPU cluster using approximately 2000 Intel Xeon cores running at 2.1Ghz simultaneously.
We utilized approximately 2.4 CPU years to collect the small control experiments data used for this study, with all post-processing, analysis, and plotting done locally on a laptop.
The large demonstration on the DMC-\shbabb{} required approximately 1.3 GPU years of compute.

\section{Additional Results}\label{app:additional-results}
% --> A bit more detail of the results included in the main body
In this section, we provide additional results of experiments run in Section~\ref{sec:evaluation}.
For these results, we use the same experimental setup as in Section~\ref{sec:evaluation}, namely we form a dataset of 250 runs of every \hyper{} setting, environment, and algorithm tuple.
From this extensive dataset, we use bootstrap resampling to simulate experimental trials using the \shbabb{}.
We start by investigating a slice of the sample distributions from which we perform resampling, then we provide additional results demonstrating the high variance of conclusions drawn from tuning on a subset of environments.

\subsection{Distribution of selected \hypers{}.}\label{app_hypersensitivity}
\begin{figure}[H]
  \centering

  \begin{minipage}[c]{0.5\textwidth}
    \includegraphics[width=\textwidth]{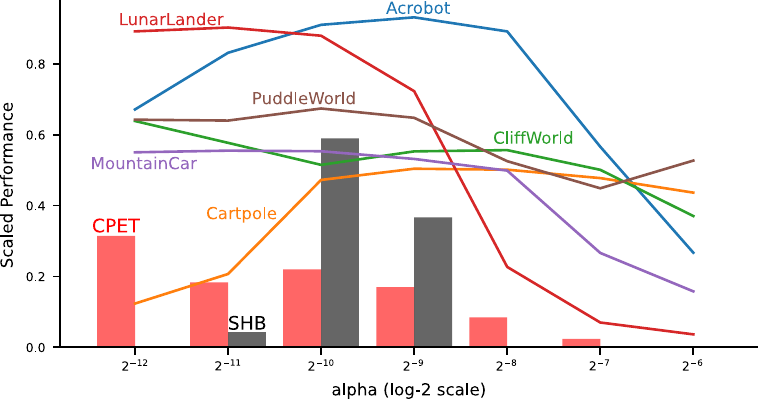}
  \end{minipage}\hfill
  \begin{minipage}[c]{0.5\textwidth}
    \caption{\label{fig:alpha-dist-shb}
      Bars represent the distribution of selected stepsizes using conventional per-environment tuning (\textbf{red}) or when using the \shbabb{} (\textbf{black}).
      Lines show the sensitivity curves for each environment.
      Confidence regions around the sensitivity curves are negligibly small and are not visible when plotted.
    }
  \end{minipage}
  \vspace{-0.6cm}
\end{figure}
We show that the distribution of selected \hypers{} with the \shbabb{} is narrow and consistent over simulated experiments, unlike parameters chosen independently for each environment.
Because conclusions are often drawn by aggregating results over environments---either formally as in the \shbabb{} or informally by counting wins---reporting results over a consistent and narrow distribution of \hypers{} leads towards lower variance claims and greater reproducibility.

Figure~\ref{fig:alpha-dist-shb} demonstrates the wide range of stepsizes used to draw conclusions using repeated applications of the conventional per-environment tuning approach and the relatively narrow range used by repeated applications of the \shbabb{}.
DQN does not have a consistently good stepsize setting that solves every environment, or even a majority of environments.
Several sensitivity curves have fairly opposite performance for a given stepsize, demonstrating the difficulty in picking a single stepsize with which to evaluate DQN.\@

Previous works generally investigate sensitivity over \hypers{} on each environment individually.
This within-environment investigation empirically shows the deviation in performance of an algorithm if different settings of a \hyper{} were used, indicating the difficulty of selecting \hypers{} for just that environment.
Our goal is slightly different.
We seek to measure the difficulty of selecting \hypers{} across multiple environments.
Consider DQN's sensitivities in Figure~\ref{fig:alpha-dist-shb}.
By looking at Acrobot, Cartpole, and Mountain Car---a commonly used suite of classic control environments---we might conclude that DQN is across-environment insensitive because it is simultaneously within-environment insensitive for these environment.
However, expanding our investigation by including Puddle World we see again that DQN is within-environment insensitive, but across-environment highly sensitive; Cartpole and Puddle World have very few overlapping good stepsizes.
Adding the Lunar Lander environment and it is clear that DQN is not within-environment insensitive, and as such is highly unlikely to be across-environment insensitive as well.

\subsection{Tuning on a subset}
\begin{figure}[H]
  \centering
  \includegraphics[width=\textwidth]{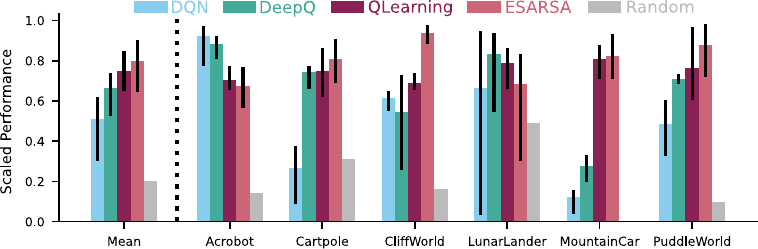}
  \vspace{0.01cm}
  \caption{\label{fig:oos-2d-3r}
    Outcome of the \shbabb{} when using \textbf{two} randomly selected domains to tune \hypers{}, then evaluating \hypers{} on all six domains.
    Each of the 10k simulated experiments use three seeds to select \hypers{}, then 250 seeds to evaluate performance.
  }
\end{figure}

In Section~\ref{sec:evaluation}, we investigate the impact of using a subset of environments to tune \hypers{} while reporting results on the full set of environments.
We demonstrated the high variance of conclusions using two randomly selected domains for each simulated experiment and using 100 random seeds to pick \hypers{}.
In Figure~\ref{fig:oos-2d-3r}, we demonstrate even greater variance in conclusions when using only three seeds to pick \hypers{}; using a consistent number of seeds as the rest of our prior evaluation.
In this setting, the correct ordering of algorithms is detected in only approximately 40\% of experiments, with distinguishing between QLearning and ESARSA providing the largest source of error.
Notice also that the variance in Lunar Lander for DQN is such that the 95\% confidence interval about the mean states that the true mean is 95\% likely to lie in the interval $[0.05, 0.98]$ where the performance metric is bounded between 0 and 1.
In other words, evaluating the performance of DQN on Lunar Lander using this experimental design is effectively useless.

\begin{figure}[H]
  \centering
  \includegraphics[width=\textwidth]{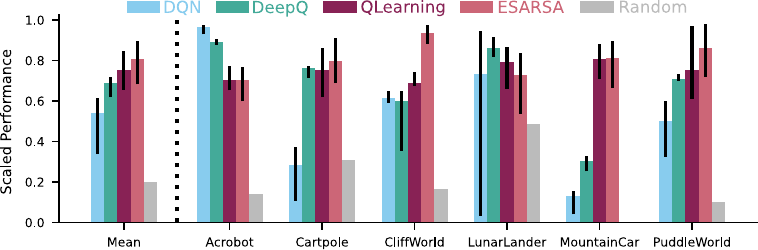}
  \vspace{0.01cm}
  \caption{\label{fig:oos-3d-100r}
    Outcome of the \shbabb{} when using \textbf{three} randomly selected domains to tune \hypers{}.
    Each of the 10k simulated experiments use 100 seeds to select \hypers{}, then 250 seeds to evaluate performance.
  }
\end{figure}

\begin{figure}[H]
  \centering
  \includegraphics[width=\textwidth]{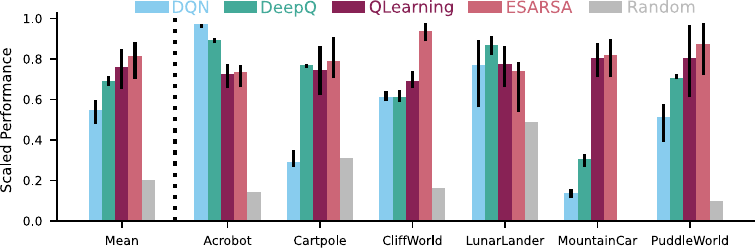}
  \vspace{0.01cm}
  \caption{\label{fig:oos-4d-100r}
    Outcome of the \shbabb{} when using \textbf{four} randomly selected domains to tune \hypers{}.
    Each of the 10k simulated experiments use 100 seeds to select \hypers{}, then 250 seeds to evaluate performance.
  }
\end{figure}

We continue our investigation of using a subset of environments to select \hypers{} in Figures~\ref{fig:oos-3d-100r} and~\ref{fig:oos-4d-100r} where we use three of six environments and four of six environments respectively.
In both figures we use 100 random seeds to select \hypers{}, consistent with Figure~\ref{fig:oos-shb} in Section~\ref{sec:evaluation}.
In Figure~\ref{fig:oos-3d-100r}, the variance in overall conclusions is notably smaller than when using two of six environments to select \hypers{}.
This experimental design allows distinguishing between DQN and DeepQ reliably, but still fails to distinguishing the performance of QLearning and ESARSA.
The variance of DQN on Lunar Lander still provides a comically large confidence region.

Figure~\ref{fig:oos-4d-100r} uses four of six environments with 100 random seeds for each \hyper{} setting to select \hypers{}.
The confidence region around DQN and DeepQ indicates a clear and meaningful ordering in the performance of the deep RL algorithms across these environments.
Still QLearning and ESARSA remain indistinguishable.
The variance on individual environments is sensible, allowing some conclusions to be drawn with confidence especially when comparing DeepQ and DQN.
We point out that the computational savings of using four of six environments is entirely negated by the number of random seeds required to select \hypers{} in this experimental design, calling to question the utility of subselecting environments.
This suggests that this experimental design is likely not yet appropriate for use in RL due to high variance in conclusions---at least until future algorithm development yields algorithms with significantly less across-environment sensitivity.

\subsection{Performance distributions}

\begin{figure}[H]
  \centering
  \begin{subfigure}{.32\textwidth}
    \caption*{Cartpole}
    \includegraphics[width=\textwidth]{plots/alpha-dist/cartpole-DQN_500_32}
  \end{subfigure}
  \begin{subfigure}{.2648\textwidth}
    \caption*{Acrobot}
    \includegraphics[width=\textwidth]{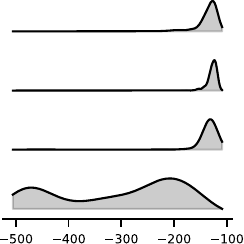}
  \end{subfigure}
  \begin{subfigure}{.253\textwidth}
    \caption*{CliffWorld}
    \includegraphics[width=\textwidth]{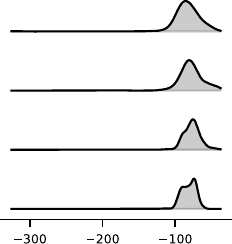}
  \end{subfigure}

  \vspace{0.75cm}
  \begin{subfigure}{.32\textwidth}
    \includegraphics[width=\textwidth]{plots/alpha-dist/puddleworld-DQN_500_32}
    \caption*{PuddleWorld}
  \end{subfigure}
  \begin{subfigure}{.2555\textwidth}
    \includegraphics[width=\textwidth]{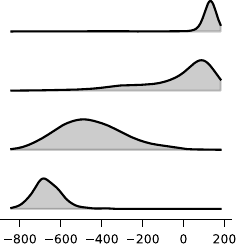}
    \caption*{Lunar Lander}
  \end{subfigure}
  \begin{subfigure}{.2535\textwidth}
    \includegraphics[width=\textwidth]{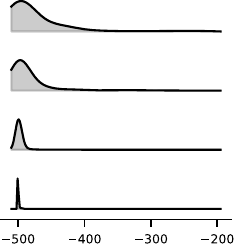}
    \caption*{Mountain Car}
  \end{subfigure}

  \vspace{0.5cm}
  \caption{\label{fig:all-stepsize-dists}
    Various slices of $\CP{\Perf}{\env, \param}$ for all $\env \in \Envs$ and a subset of $\Params$ for the DQN algorithm.
    Every distribution is estimated with a Gaussian kernel density estimator and 250 samples.
    The supports for the distributions are computed by finding the absolute min and max run for the visualized \hyper{} settings.
    This means, for instance, that at least one run shown on Mountain Car achieved a performance of approximately -200 return, but was such a low probability event it is not visible in these plots.
  }
\end{figure}

Figure~\ref{fig:all-stepsize-dists} demonstrates that the shapes of the performance distributions are highly inconsistent across environment and choice of stepsize parameter.
It is clear that assuming normality of the data is in general impossible, even for these simple algorithms and small domains.
Experiments that use only a small number of random seeds---especially when maximizing over repeated trials, or cherry-picking over completed results---are highly unlikely to capture the bimodality and skew present in many of these distributions.
Consider, for instance, the Lunar Lander environment with DQN.
Using a small number of random seeds---for instance three---it is highly unlikely that the long-tail of the distribution for stepsize $\alpha=2^{-9}$ is accurately captured.
Instead, the most likely outcome is that the mean of the high-performing mode is reported, ignoring the instability of the DQN algorithm.

\subsection{Results when using worst-case performance across environments}\label{app_worstcase}
\begin{figure}[H]
  \centering

  \includegraphics[width=0.8\textwidth]{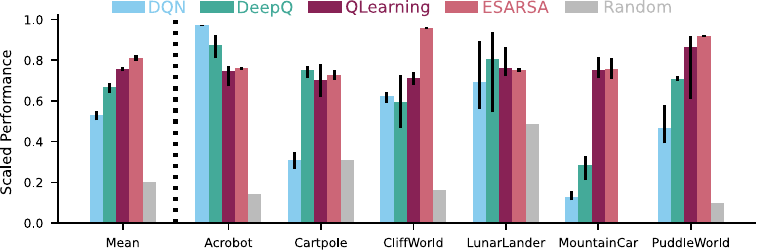}
  \vspace{0.02cm}
  \caption{\label{fig:percentile}
    Performance of each algorithm over 10k bootstrap samples, where sample means are computed with 3 runs.
    \Hypers{} are selected for maximizing performance of the worst-case environment.
    Error bars show 95\% bootstrap confidence intervals.
  }
\end{figure}

The prior evaluations of the \shbabb{} all estimated $\CE{\normalizeE{\Perf}}{\shbparam}$ with \hypers{} selected to maximize this expectation.
However, we could instead select \hypers{} according to other statistics, for instance those with best performance on the worst-case environment, $\max_{\param} \min_{\env} \CE{\normalizeE{\Perf}}{\env, \param}$.
We demonstrate in Figure~\ref{fig:percentile} the outcome of the \shbabb{} when $\shbparam$ is selected according to maximizing performance on the worst-case environment.
Note the nested optimization means the environment chosen may be different per algorithm and even per \hyper{} setting.

In general, the variability of conclusions made over 10k simulated applications of the \shbabb{} is much smaller than when estimating the mean over environments, as shown in Figure~\ref{fig:percentile}.
We still see high variance in the Lunar Lander environment for the two DRL algorithms, suggesting still high sensitivity to the chosen \hypers{}.
Because the worst-case environment for DQN is Cartpole for many choices of \hyper{}, likewise Mountain Car for DeepQ, it is likely that these environments largely dictate the choice of \hyper{} without regard for performance on Lunar Lander.
Under this worst-case benchmark, algorithm development improving the performance of DQN on Cartpole would be highly rewarded while development slightly improving its performance on Acrobot would have no effect.
This is unlike conventional benchmarks where minute improvements on already well-solved problems are rewarded similarly to large improvements on challenging problems.

\subsection{DMControl demonstration}\label{app:dmcontrol-results}

\begin{figure}[t]
  \centering

  \includegraphics[width=0.75\textwidth]{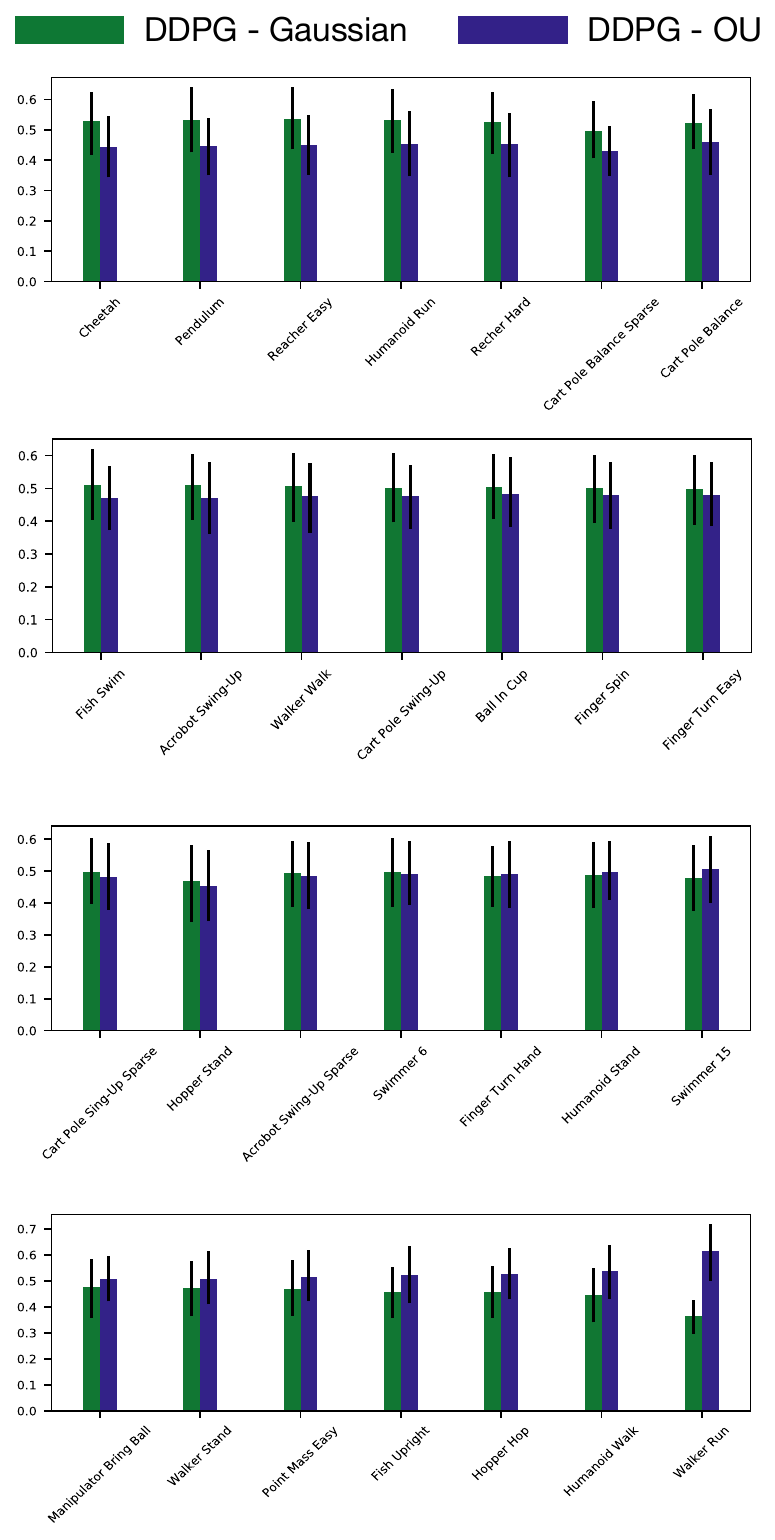}
  \caption{\label{fig:dmcontrol}
    Per-environment performance differences for every environment in the DMControl suite.
    Error bars show 95\% confidence intervals using 1k bootstrap samples.
    Performance is averaged over 30 independent runs.
  }
\end{figure}

Figure~\ref{fig:dmcontrol} shows the per-environment performance for DDPG using both OU noise and Gaussian noise evaluated using 30 runs of the $\shbparam$ setting.
In most cases, the performance of each exploration method was not statistically significantly different.
Although there is a large difference in the WalkerRun environment, we point out this may be due to the \shbabb{} trading-off performance on other environments in order to pick a single \hyper{} setting; without explicit per-environment experimentation this remains unclear.

\section{Further Experimental Details}\label{app:experiment-details}

In this section we include all experimental details used in Section~\ref{sec:evaluation}, descriptions of the environments used are included in Appendix~\ref{app:small-control-environments}, and details about the DMControl demonstration from Section~\ref{sec:big-demo} are included in Appendix~\ref{app:dmcontrol}.

The tile-coded agents both learn directly from the most recent observations without the use of a replay buffer.
We use stochastic gradient descent to optimize the agents, with stepsizes scaled by the number of active tiles in the representation---equal to the number of tilings used by the tile-coder except in the special case of Lunar Lander.
Like their deep RL counterparts, we use $\epsilon$-greedy policies to train the tile-coding agents with $\epsilon=0.1$ for all experiments.

\subsection{Small control environments}\label{app:small-control-environments}
In this section, we describe the environments used to evaluate the \shbabb{} in Section~\ref{sec:evaluation}.
Our goal in environment selection was to highlight differences between the demonstrative algorithms, while simultaneously using small enough environments to feasibly collect an extensive dataset to justify our experiment design.
Because all of our demonstrative algorithms use $\epsilon$-greedy action selection as their sole form of exploration, we avoid environments where exploration is a particular challenge as this would not help in distinguishing between algorithms.

For the Acrobot environment \citep{sutton1996generalization}, we use the implementation from \citet{brockman2016openai}.
Acrobot has a medium-sized observation dimension with six observable values, making feature representation challenging for tile-coding agents.
Similarly, we include the Lunar Lander environment \citep{brockman2016openai} as its observation dimension is too large for tile-coding to successfully generate a useful representation.
Lunar Lander additionally has a highly shaped reward function, making the learning dynamics very different from all other included environments.

For the Cartpole \citep{barto1983neuronlike} and Mountain Car \citep{moore1990efficient} environments, we likewise use the implementation from \citet{brockman2016openai}.
Both environments have a small observation dimension, making the feature representation amenable to tile-coding.
Prior results have suggested a stark difference in performance between DQN and DeepQ on these environments, suggesting their utility in distinguishing between algorithms.

Lastly, CliffWorld \citep{sutton2018reinforcement} and PuddleWorld \citep{sutton1996generalization} are both two dimensional gridworlds.
The small observation dimension is easier for tile-coding agents to represent and presents a challenge for neural network based agents.
CliffWorld is commonly used to showcase large differences between on-policy and off-policy algorithms \citep{sutton2018reinforcement}, making it a good choice for differentiating between the three Q-learning based agents and ESARSA.\@
Additionally, the sudden large negative reward obtained from falling off the cliff could cause high variance updates for mean squared based algorithms, suggesting a slight advantage for DQN.\@
PuddleWorld uses a dense reward function with shaping, making it similar to Lunar Lander.

\subsection{Details about DM Control demonstration}\label{app:dmcontrol}
% --> Details about the DMControl experiments
The demonstration in Section~\ref{sec:big-demo} was generated using the Acme codebase of RL algorithms \citep{hoffman2020acme}.
We reuse as much code from Acme as possible to maintain similarity in empirical setup and computational cost with prior works coming from this lab, e.g \citet{tassa2018deepmind,lillicrap2016continuous,barth-maron2018distributed}.
We use the default \hypers{} and network architectures for all experiments and environments as in Acme, except for those which we swept.
We used 3 random seeds for each environment and \hyper{} setting to select \hypers{} according to the \shbabb{}, then we perform an additional 30 runs to evaluate $\shbparam$.

For the \hyper{} sweep, we evaluated stepsize $\alpha \in \{10^{-4}, 10^{-3}, 10^{-2}\}$ for the critic and $\eta \in \{10^0, 10^{-1}, 10^{-2}\}$ where $\beta = \eta\alpha$ and $\beta$ is the stepsize for the actor network.
We use the ADAM optimizer with default parameters.
We additionally swept over target network types, using either Polyak averaging with moving average parameter $\beta_{tn} = 0.001$ or a hard refresh every 100 steps.
Finally, we swept the standard deviation of the exploration noise $\sigma \in \{0.05, 0.1\}$---a slight deviation in ranges tested by previous works as we noticed $\sigma > 0.1$ was rarely a good choice on most environments, but $\sigma < 0.1$ was often required to obtain better than random performance on several environments (e.g. Acrobot).

All experiments run for a maximum of 300k learning steps and use an infinite replay buffer.
On every environment interaction after the first 1000 steps, the DDPG agent made a mini-batch update using a mini-batch size of 64.
To maintain consistency with prior works, a soft-termination occurs after 1000 steps in an episode.

Finally, we include the \hypers{} selected by the DMC-\shbabb{}.
Both DDPG and DDPG-OU selected the same \hypers{} when using the DMC-\shbabb{}.
Defaults taken from the Acme codebase \citep{hoffman2020acme}, as was the code implementation.

\begin{minipage}[c]{\textwidth}
\lstset{
  basicstyle=\footnotesize\ttfamily,
  showspaces=false,
  numberstyle=\footnotesize,
  numbersep=9pt,
  tabsize=2,
  breaklines=true,
  showtabs=false,
}
\begin{lstlisting}
{
"max_steps": 300000,
  "metaParameters": {
    "actor_stepsize_scale": 1.0,
    "critic_stepsize": 1e-4,
    "discount": 0.99,
    "target_update": 0.001,

    "buffer_size": "infinite",
    "min_replay_size": 1000,
    "steps_per_update": 1,
    "batch": 64,

    "n_step": 1,
    "sigma": 0.1,
    "theta": 0.15,
    "mu": 0.0,
    "clipping": false,

    "obs_weights": [[400, 400]],
    "policy_weights": [[300, 200]],
    "critic_weights": [[400, 300]],
  }
}
\end{lstlisting}

\end{minipage}

\end{document}